\documentclass[a4, 10 pt, twocolumn]{article}  

\usepackage{times}
\usepackage{multicol}
\usepackage[bookmarks=true]{hyperref}

\usepackage{geometry}
\usepackage{authblk}
\usepackage[pdftex]{graphicx} 
\usepackage{titlesec}
\usepackage[labelfont=bf]{caption}
\usepackage{enumitem}
\usepackage{hyperref}
\usepackage{helvet}
\usepackage{etoolbox}
\usepackage{graphics}

\DeclareCaptionFormat{myformat}{\fontsize{9}{9}\selectfont#1#2#3}
\titleformat{\section}
  {\normalfont\fontsize{10}{0}\bfseries}{\thesection}{0em}{}
\titlespacing\section{0pt}{12pt }{3pt }
\makeatletter
\patchcmd{\@maketitle}{  \@title}{\fontsize{14}{0}\selectfont\sffamily \bfseries \@title}{}{}
\makeatother

\usepackage{hyperref}   
\usepackage{graphicx}
\usepackage{setspace}
\usepackage{array}
\usepackage{longtable}
\usepackage{multirow}
\usepackage{hhline}
\usepackage{rotating}
\usepackage{amssymb, amsmath, mathrsfs}

\usepackage{color}
\graphicspath{{./images/}}
\DeclareGraphicsExtensions{.pdf,.png,.jpg}
\usepackage{algorithm}
\usepackage{algorithmic}
\usepackage{subfigure}
\usepackage{tabularx}
\usepackage{multirow}
\usepackage{kotex}
\usepackage{comment}
\usepackage{xcolor, soul}
\usepackage{pifont}

\specialcomment{keyword-box}{\begingroup\color{gray}}{\endgroup}

\definecolor{orange}{rgb}{1,0.2,0}

\newcommand{\captionfonts}{\small}

\makeatletter  
\long\def\@makecaption#1#2{%
  \vskip\abovecaptionskip
  \sbox\@tempboxa{{\captionfonts #1: #2}}%
  \ifdim \wd\@tempboxa >\hsize
    {\captionfonts #1: #2\par}
  \else
    \hbox to\hsize{\hfil\box\@tempboxa\hfil}%
  \fi
  \vskip\belowcaptionskip}
\makeatother   

\newcounter{tecounter}
\newenvironment{tightenumerate}
{
	\begin{list}{\arabic{tecounter}\addtocounter{tecounter}{1}.}{%
			\setcounter{tecounter}{1}
			\setlength{\leftmargin}{12pt}
			\setlength{\topsep}{1pt}
			\setlength{\partopsep}{0pt}
			\setlength{\itemsep}{2pt}
			\setlength\labelwidth{7pt}}
		\ignorespaces}
	{\unskip\end{list}
}

\makeatletter
\let\NAT@parse\undefined
\makeatother
\usepackage[numbers,compress]{natbib}
\bibpunct{[}{]}{,}{n}{}{;}
\title{\vspace{-30pt} \LARGE Design and validation of zero-slack separable manipulator for Intracardiac Echocardiography
\vspace{-10pt}
}

\author[1,2]{\large Christian DeBuys}
\author[1]{\large Florin Ghesu}
\author[2]{\large Reza Langari}
\author[1]{\large Young-Ho Kim \vspace{-8pt}}
\affil[1]{\large Siemens Healthineers, Digital Technology \& Innovation, Princeton, NJ, USA \vspace{-0pt}}
\affil[2]{\large Texas A\&M University, College Station, TX, USA \vspace{-3pt}}
\affil[]{\footnotesize  \{christian.debuys, florin.ghesu, young-ho.kim\}@siemens-healthineers.com, $^2$ \{cldebuys, rlangari\}@tamu.edu \vspace{-15pt}}

\begin{document}
\date{}
\maketitle
\thispagestyle{empty}
\pagestyle{empty}

\section*{INTRODUCTION}\vspace{-5pt}
Intracardiac echocardiography (ICE) catheter, known to have a strong ability to visualize cardiac structures and blood flow from within the heart, is now being favorably used in cardiac catheterization and electrophysiology as an advanced imaging approach. 
However, clinicians require substantial training and experience to become comfortable with steering the catheter to localize and measure the area of treatment to watch for complications while device catheters are deployed in another access.
Thus, it is reasonable that a robotic-assist system to hold and actively manipulate the ICE catheter could ease the workload of the physician. 

There exist commercially-available robotic systems \citep{stereotaxis20} and research prototypes \citep{lee21ice,kim2022automated} for ICE catheter manipulation. 
They all use existing commercially available ICE catheters ({\em e.g.} ACUSON AcuNav ICE catheter family, Siemens Healthineers) based on multiple tendon-sheath mechanism (TSM). To motorize the existing TSM-based ICE catheter, the actuators interface with the outer handle knobs to manipulate four internal tendons.
However, in practice, the actuators are located at a sterile, safe place far away from the ICE handle\,\citep{kim2022automated}. Thus, to interface with knobs, there exist multiple coupled gear structures between two, leading to a highly nonlinear behavior ({\em e.g.} various slack, elasticity) alongside hysteresis phenomena in TSM\,\citep{lee21ice}.


Since ICE catheters are designed for single use, the expensive actuators need to be located in a safe place so as to be reusable. Moreover, these actuators should interface as directly as possible with the tendons for accurate tip controls. In this paper, we introduce a separable ICE catheter robot with four tendon actuation: one part reusable and another disposable. Moreover, we propose a practical model and calibration method for our proposed mechanism so that four tendons are actuated simultaneously allowing for precise tip control and mitigating issues with conventional devices such as dead-zone and hysteresis with simple linear compensation. We consider an open-loop controller since many available ICE catheters are used without position-tracking sensors at the tip due to costs and single use.




\begin{figure}[t!]
	\centering
	\includegraphics[scale= 0.33]{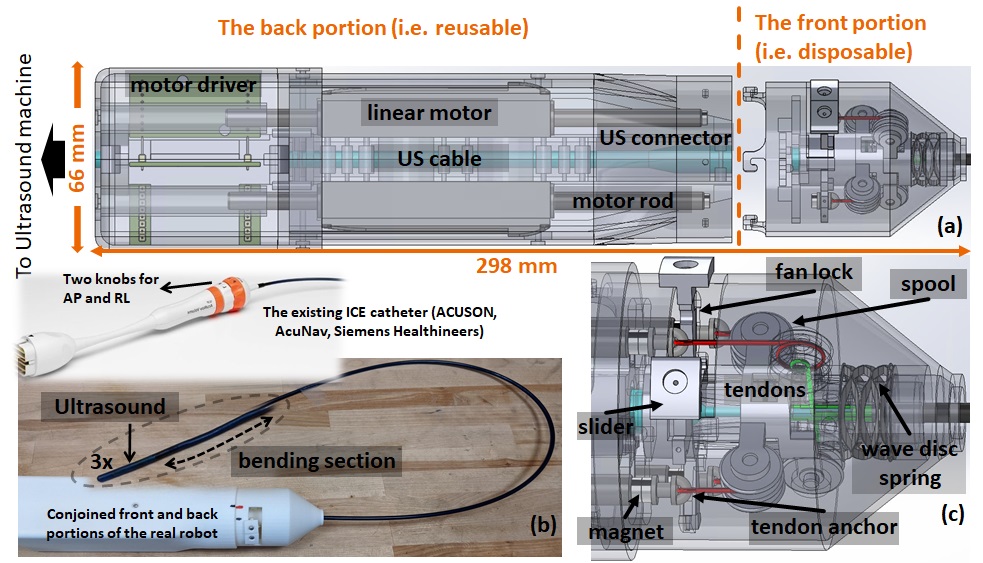}
	\vspace*{-15pt}
	\caption{An overview of the catheter robot.\label{fig:robot}}
	\vspace*{-15pt}
\end{figure}

\vspace{-8pt}
\section*{MATERIALS AND METHODS}\vspace{-5pt}
Figure\,\ref{fig:robot} explains an overview of the new concept design of ICE catheter robot. 
The proposed robotic system consists of two parts: {\em ``The back portion"} contains four Faulhaber linear motors (LM 1483-080-11-C) with motor drivers mounted to a plastic core. The plastic core contains a channel along its central axis for the ultrasound (US) cable and its connector, which are shown in blue in Fig.\,\ref{fig:robot}(a). 
The interface from one catheter tendon in {\em ``the front portion"} shown in Fig.\,\ref{fig:robot}(c) to its respective linear motor is as follows: 1) the tendon (green) from the catheter is bent 90 degrees around a low-friction roller and is wrapped around and fastened to the small radius of the spool, 2) a separate tendon (red) is wrapped around and fastened to the large radius of the spool and then attached to the tendon anchor, and 3) the tendon anchor is held in place by the fan lock until the catheter is clipped together and the motor rod has connected to the tendon anchor via magnetic force. The spools serve to increase the pulling force of the motors with a 3:1 ratio.

The clipping system functions like closing a child-proof medicine bottle: the front portion is pushed into the body until it makes contact, pushed slightly farther and twisted, then released. 
After releasing, the front is locked, the motors move their rods toward the front until their magnets make contact with the magnets on the tendon anchors, the sliders are pushed away from each other to manually open the fan-lock, and the motors are free to pull on the tendons.
The robot has two degrees of freedom, anterior/posterior (AP) curvature $\kappa_{x}$ and right/left (RL) curvature $\kappa_{y}$, and four linear actuators, making this a redundant system with four inputs and two outputs. Whole body translation/rotation are not handled in this paper. A redundant control scheme is used with a constant curvature assumption to resolve and take advantage of this redundancy, in which the energy of the control input is minimized while respecting feasibility constraints (such as requiring a minimum tension for all tendons) as in \cite{platform}. 
We define the robot's configuration $q = [\kappa_{x}, \kappa_{y}]^T$, where curvature is chosen rather than bending angle to obtain linear kinematic and static equations as in \cite{camarillo}, and axial compression is considered negligible.
We assume quasi-static motion of the catheter governed by the following moment balance equation:
\vspace{-10pt}
{\small
\begin{equation}
\begin{bmatrix} K_b~~0 \\ ~0~~K_b \end{bmatrix} \begin{bmatrix} \kappa_x \\ \kappa_y \end{bmatrix} = \begin{bmatrix} -d_{y1}~ -d_{y2}~\cdots -d_{yn} \\ ~~~d_{x1}~~~~d_{x2}~ \cdots ~~~ d_{xn} \end{bmatrix} \begin{bmatrix} T_1 \\ T_2 \\ \vdots \\ T_n \end{bmatrix}\vspace*{-5pt}
\label{eq_moment_incompressible}
\end{equation}
}
where $K_b$ is bending stiffness of the bending section, $d_{xi}$ and $d_{yi}$ are the $i^{th}$ tendon's $x$ and $y$ coordinates relative to the central axis, $T_i$ is the tension in the $i^{th}$ tendon, and $n=4$ for our device.
We can write Eq.\eqref{eq_moment_incompressible} in matrix form as $\textbf{K} \textbf{q} = \textbf{D} \tau$ in matrix form.
To convert our control input from actuator forces $\tau$ to actuator displacements $\textbf{y}$, we consider the conservation of strain equation in \cite{camarillo}.
\vspace*{-5pt}
{\small
\begin{equation}
\textbf{y} = (\textbf{D}^T \textbf{L}_0 \textbf{K}^{-1} \textbf{D} + \textbf{L}_t \textbf{K}_t^{-1}) \tau = \textbf{G} \tau
\label{eq_strain_tau}\vspace*{-5pt}
\end{equation}
}
$\textbf{L}_0$, $\textbf{L}_t$, and $\textbf{K}_t$ are diagonal matrices containing the undeformed bending section length $l_0$, the unstretched tendon lengths $l_{t,i}$, and the length-normalized tendon stiffnesses $k_{t,i}$, respectively. $\textbf{G}$ is the compliance matrix. 
Combining Eq.\,\eqref{eq_moment_incompressible} and \eqref{eq_strain_tau} we obtain the forward kinematics.
\vspace*{-5pt}
{\small
\begin{equation}
\textbf{q} = \textbf{K}^{-1} \textbf{D} \textbf{G}^{-1} \textbf{y}
\label{kinematics}
\vspace*{-5pt}
\end{equation}
}
$\textbf{K}$ is diagonal and thus invertible, and $\textbf{G}$ is invertible for any configuration of non-slack tendons. 
All parameters in the equations must be determined empirically. $\textbf{L}_0$, $\textbf{L}_t$, and $\textbf{K}_t$ can be measured directly; $\textbf{K}$ and $\textbf{D}$ can be approximated by the following parameter identification procedure:
\begin{tightenumerate}
{\small
  \item Make initial guess for parameters (K,D). \vspace{-8pt}
  \item Input motor position trajectories which should achieve desired bending angle as calculated from inverse kinematics and redundant control input. \vspace{-8pt}  
  \item Update parameter guesses based on difference between desired and measured bending angle. \vspace{-8pt} 
  \item Repeat steps 2 and 3 until parameter values converge.}
\end{tightenumerate}


The desired bending angle was changed to desired curvature and then converted to the desired motor position inputs using the parameter estimates and by inverting Eq.\,\eqref{kinematics}. The inverse kinematics were solved with an energy minimizing redundant control input which included minimum tension constraints to prevent slack \cite{platform}. 

The proposed methods were validated with three trials on a test bed by following a desired sinusoidal output for bending angle in the AP/LR direction. For each trial, the robot followed two cycles of the sinusoid, and the ground truth output angle was measured with an EM sensor, which was only used to evaluate performance.
Tip position and bending angle errors were described as mean absolute error (MAE) and standard deviation (StD).
Performance was compared without and with simple hysteresis compensation. 
The simple hysteresis compensation involved adjusting the input angle by a constant $\pm$ 10 degrees depending on the direction of motion. The backlash width (about $20^\circ$) was obtained based on the unique physical properties of the catheter\,\citep{lee21ice}.



\section*{RESULTS} 
\begin{figure}[t!]
	\centering
	\includegraphics[scale= 0.38]{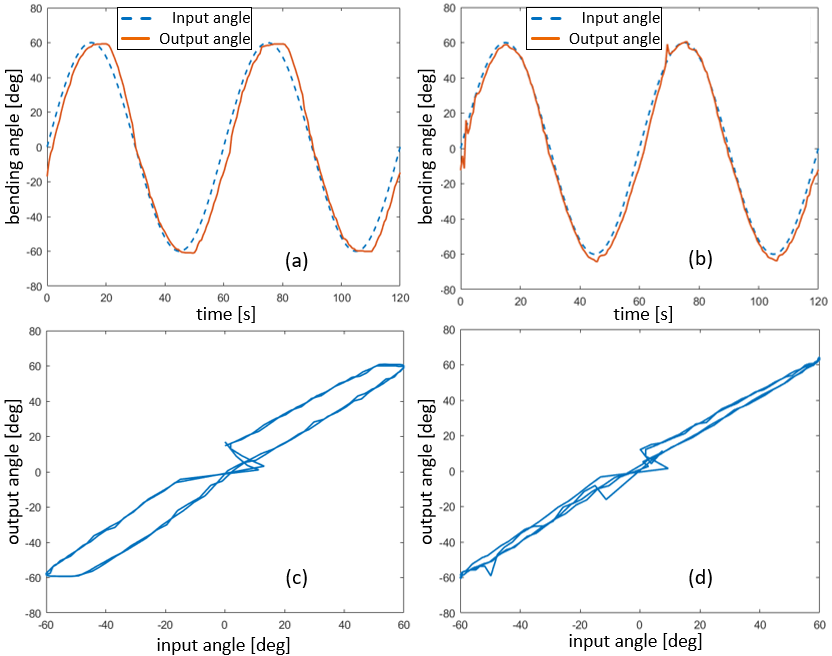}
	\vspace*{-5pt}
	\caption{One result for AP bending: (a) Time versus output angle without compensation (b) Time versus output angle with hysteresis compensation (c)(d) The input angle versus the output angle without/with hysteresis compensation }
	\label{fig:result}
	\vspace*{-15pt}
\end{figure}

\vspace{-5pt}
Figure \ref{fig:result} shows one example of bending angle for without and with our simple hysteresis compensation. 
The control of all tendons simultaneously allowed the removal of tendon slack and thereby removed the deadzone presented in conventional catheters. Later, the performance is improved with a simple hysteresis compensator.
The overall performance evaluation is described in Table\,\ref{Tab:error}. The tip pose errors (MAE) are reported as $4.8~mm$ and $6.1^\circ$. However, the position and orientation errors are improved $31\%$ and $47\%$ with a simple hysteresis compensation, respectively.

\begin{table}[h!]
\centering
\vspace{-15pt}
\caption{Mean Absolute Error (MAE) and Standard Deviation (StD) for Tip Position and Bending Angle}
\scalebox{0.7}{
\begin{tabular}{||c c c c c||} 
 \hline
  & {\small Position [mm]} & {\small StD} & {\small Angle [deg]} & {\small StD} \\ [0.5ex] 
 \hline\hline
  {\small no comp.} & 4.80 & 1.97 & 6.11	& 4.03 \\ 
 \hline
 {\small comp.} & 3.31 & 2.02 & 3.26 & 2.90 \\
 \hline
 {\small \% reduction} & 31.0  &   & 46.6 &  \\ 
 \hline
\end{tabular}
}
 \vspace*{-15pt}
\label{Tab:error}
\end{table}

\section*{DISCUSSION}\vspace{-5pt}
Our proposed design and models are validated in a simple scenario. As a future work, we plan to investigate more sophisticated controls ({\em e.g.,} a predictive or learning-based) for complicated scenarios with a systematic evaluation.




\vspace*{-5pt}
\section*{DISCLAIMER}\vspace{-5pt}
{
The concepts and information presented in this paper are based on research results that are not commercially available. 
%




}
\vspace*{-5pt}
\section*{REFERENCES}\vspace{-5pt}
\renewcommand{\section}[2]{}
{
\small
\bibliography{references_hamlyn}
}

\end{document}